\documentclass{article} 
\usepackage{iclr2022_conference,times}

\usepackage{amsmath,amsfonts,bm}

\def\eqref#1{equation~\ref{#1}}

\def\1{\bm{1}}

\DeclareMathAlphabet{\mathsfit}{\encodingdefault}{\sfdefault}{m}{sl}
\SetMathAlphabet{\mathsfit}{bold}{\encodingdefault}{\sfdefault}{bx}{n}

\newcommand{\R}{\mathbb{R}}

\DeclareMathOperator*{\argmax}{arg\,max}

\usepackage{paralist, tabularx}

\usepackage[utf8]{inputenc} 
\usepackage[T1]{fontenc}    
\usepackage{hyperref}       
\usepackage{url}            
\usepackage{booktabs}       
\usepackage{amsfonts}       
\usepackage{nicefrac}       
\usepackage{microtype}      
\usepackage{xcolor}         

\usepackage{amsmath}
\usepackage{multirow}
\usepackage{threeparttable}
\usepackage{tabularx}
\usepackage{paralist}

\usepackage{bm}
\usepackage{bbm}

\usepackage{amsthm}
\usepackage{booktabs}       
\usepackage{nicefrac}       
\usepackage{microtype}      

\usepackage{wrapfig}

\usepackage{siunitx}

\usepackage{subfig}
\usepackage{graphicx}

\usepackage{xspace}
\newcommand{\model}{\textsc{ZESRec}\xspace}
\newcommand{\nop}[1]{}

\def\l{{\bf l}}

\def\I{{\bf I}}

\def\R{{\bf R}}
\def\X{{\bf X}}

\def\S{{\bf S}}

\def\x{{\bf x}}

\def\m{{\bf m}}
\def\n{{\bf n}}
\def\U{{\bf U}}
\def\u{{\bf u}}
\def\V{{\bf V}}
\def\v{{\bf v}}

\def\0{{\bf 0}}
\def\1{{\bf 1}}

\def\ep{\mbox{\boldmath$\epsilon$\unboldmath}}

\def\xii{\mbox{\boldmath$\xi$\unboldmath}}

\def\argmax{\mathop{\rm argmax}}

\newcommand{\secref}[1]{Sec.~\ref{#1}}
\newcommand{\figref}[1]{Fig.~\ref{#1}}

\newcommand{\eqnref}[1]{Eqn.~\ref{#1}}

\author{Antiquus S.~Hippocampus, Natalia Cerebro \& Amelie P. Amygdale \thanks{ Use footnote for providing further information
about author (webpage, alternative address)---\emph{not} for acknowledging
funding agencies.  Funding acknowledgements go at the end of the paper.} \\
Department of Computer Science\\
Cranberry-Lemon University\\
Pittsburgh, PA 15213, USA \\
\texttt{\{hippo,brain,jen\}@cs.cranberry-lemon.edu} \\
\And
Ji Q. Ren \& Yevgeny LeNet \\
Department of Computational Neuroscience \\
University of the Witwatersrand \\
Joburg, South Africa \\
\texttt{\{robot,net\}@wits.ac.za} \\
\AND
Coauthor \\
Affiliation \\
Address \\
\texttt{email}
}

\iclrfinalcopy 
\begin{document}
\setcitestyle{square}

\title{Zero-Shot Recommender Systems}

\author{Hao Ding \& Yifei Ma \& Anoop Deoras \& Yuyang Wang\\
AWS AI Labs \\
\texttt{\{haodin,yifeim,adeoras,yuyawang\}@amazon.com} \\
\And
Hao Wang \\
Rutgers University \\
\texttt{hw488@cs.rutgers.edu} \\
}

\maketitle

\begin{abstract}
Performance of recommender systems (RecSys) relies heavily on the amount of training data available. This poses a chicken-and-egg problem for early-stage products, whose amount of data, in turn, relies on the performance of their RecSys. In this paper, we explore the possibility of zero-shot learning in RecSys, to enable generalization from an old dataset to an entirely new dataset. We develop an algorithm, dubbed ZEro-Shot Recommenders (\model), that is trained on an old dataset and generalize to a new one where there are \emph{neither overlapping users nor overlapping items}, a setting that contrasts typical cross-domain RecSys that has either overlapping users or items. Different from previous methods that use categorical item indices (i.e., item ID), \model uses items' generic features, such as natural-language descriptions, product images, and videos, as their continuous indices, and therefore naturally generalizes to any unseen items. In terms of users, \model builds upon recent advances on sequential RecSys to represent users using their interactions with items, thereby generalizing to unseen users as well. We study three pairs of real-world RecSys datasets and demonstrate that \model can successfully enable recommendations in such a zero-shot setting, opening up new opportunities for resolving the chicken-and-egg problem for data-scarce startups or early-stage products.
\end{abstract}

\section{Introduction}

Many large scale e-commerce platforms (such as Etsy, Overstock, etc.) and online content platforms (such as Spotify, Overstock, Disney+, Netflix, etc) have such a large inventory of items that showcasing all of them in front of their users is simply not practical. In particular, in the online content category of businesses, it is often seen that users of their service do not have a crisp intent in mind unlike in the retail shopping experience where the users often have a {clear} intent of purchasing something. The need for personalized recommendations therefore arises from the fact that not only it is impractical to show all the items in the catalogue but often times users of such services need help discovering the next best thing — be it the new and exciting movie or be it a new music album or even a piece of merchandise that they may want to consider for future buying if not immediately.

Modern personalized recommendation models of users and items have often relied on the idea of extrapolating preferences from similar users. Different machine learning models define the notion of similarity differently. Classical bi-linear Matrix Factorization (MF) approaches model users and items via their identifiers and represent them as vectors in the latent space~\cite{MF,PMF}. Modern deep-learning-based recommender systems~\cite{CDL,GRU4Rec,HRNN}, which are also used for predicting top-$k$ items given an item, learn the {user-to-item} propensities from large amounts of training data containing many (user, item) tuples, optionally with available item content information (e.g., product descriptions) and user metadata. 

As machine learning models, the performance of RecSys relies heavily on the amount of training data available. This might be feasible for large e-commerce or content delivery websites such as Overstock and Netflix, but poses a serious chicken-and-egg problem for small startups, whose amount of data, in turn, relies on the performance of their RecSys. On the other hand, zero-shot learning promises some degree of generalization from an old dataset to an entirely new dataset. In this paper, we explore the possibility of zero-shot learning in RecSys. We develop an algorithm, dubbed ZEro-Shot Recommenders (\model), that is trained on an old dataset and generalize to a new one where there are \emph{neither overlapping users nor overlapping items}, a setting that contrasts typical cross-domain RecSys that has either overlapping users or items~\cite{DBLP:conf/sigir/Yuan0KZ20,DBLP:conf/sigir/WuYCLH020,DBLP:conf/sigir/BiSYWWX20a,li2019zero}. Naturally, generalization of RecSys to unseen users and unseen items becomes the two major challenges for developing zero-shot RecSys. 


For the first challenge on unseen users, we build on a rich body of literature on sequential recommendation models~\cite{GRU4Rec,HGRU,HRNN,SASRec,NARM,STAMP}. These models are built {with sequential structure} to encode temporal ordering of items in user’s item interaction history. 
Representing users by the items they have consumed in the past allows the model to extrapolate the preference learning to even novel users who the model did not see during training, as long as the items these unseen users have interacted with have been seen during training. However, such deep learning models encode item via its categorical item index, i.e., the item ID, and therefore fall short in predicting a likely relevant but brand-new item not previously seen during training. 

This brings us to the second challenge of developing zero-shot recommender systems, i.e., dealing with unseen items. To address this challenge, 
\model goes beyond traditional categorical item indices and uses items' {generic features such as natural-language descriptions, product images, and videos} as their continuous indices, thereby naturally generalizing to any unseen items. {Take natural-language (NL) descriptions as an example.} One can think of NL descriptions as a system of universal identifiers that indexes items from arbitrary domains. Therefore as long as one model is trained on a dataset with NL descriptions, it can generalize to a completely different dataset with a similar NL vocabulary. In \model we build on state-of-the-art pretrained NL embedding models such as BERT~\cite{devlin2018bert} to extract NL embeddings from raw NL descriptions, leading to an item ID system in the continuous space that is generalizable across arbitrary domains. For instance, in e-commerce platform, one could use items' description text; and similarly in the online content platforms, one could use movie synopsis or music track descriptions to represent an item. 

Combining the merits of sequential RecSys and the idea of universal continuous ID space, our \model successfully enables recommendation in an extreme cold-start setting, i.e., the zero-shot setting where all users and items in the target domain are unseen during training. Essentially \model tries to learn transferable user behavioral patterns in a universal continuous embedding space. For example, in the source domain, \model can learn that if users purchase snacks or drinks (e.g., `Vita Coconut Water' with a lemonade flavor) that they like, they may purchase similar snacks or drinks with different flavors (e.g., `Vita Coconut Water' with a pineapple flavor), as shown in the case study. Later in the target domain, if one user purchase `V8 Splash' with a tropical flavor, \model can recommend `V8 Splash' with a berry flavor to the user (see~\figref{fig:case_study_2} of~\secref{sec:case} for details). Such generalization is possible due to the use of the NL descriptions 
as universal identifiers, based on which \model could easily identify similar products of the same brand with different flavors. To summarize our contributions:
\begin{compactitem}
    \item We identify the problem of zero-shot recommender systems and propose \model as the first hierarchical Bayesian model for addressing this problem.
    \item We introduce the notion of universal continuous identifiers that makes recommendation in a zero-shot setting possible.
    \item We provide empirical results which show that \model can successfully recommend items in the zero-shot setting.
    \item We conduct case studies demonstrating that \model can learn interpretable user behavioral patterns that can generalize across datasets.
\end{compactitem}

\section{Related Work}

\textbf{Deep Learning for RecSys.}
Deep learning has been prevalent in modern recommender systems~\citep{RBM4CF,deepmusic,CDL,CVAE,chen2019top,fang2019deep,tang2019towards} due to its scalability and superior performance. As a pioneer work, \citep{RBM4CF} uses restricted Boltzmann machine (RBM) to perform collaborative filtering in recommender systems, however the system is a single-layer RBM. Later, \citep{CDL} and \citep{CVAE} build upon Bayesian deep learning to develop hierarchical Bayesian models that tightly integrate content information and user-item rating information, thereby significantly improving recommendation performance. After that, there are also various proposed sequential (or session-based) recommender systems~\citep{GRU4Rec,HGRU,TCN,li2017neural,liu2018stamp,wu2019session,HRNN}, GRU4Rec~\citep{GRU4Rec} was first proposed to use gated recurrent units (GRU)~\citep{GRU} for recommender systems. Since then, follow-up works such as hierarchical GRU~\citep{HGRU}, temporal convolutional networks (TCN)~\citep{TCN}, and hierarchical RNN (HRNN)~\citep{HRNN} have achieved improvement in terms of accuracy by utilizing cross-session information~\citep{HGRU}, causal convolutions~\citep{TCN}, as well as meta data and control signals~\citep{HRNN}. In this paper we build on such {sequential} RecSys and note that our \model is model agnostic, that is, it is compatible with any sequential RecSys. 

\textbf{Cross-Domain and Cold-Start RecSys.} There is a rich literature on cross-domain RecSys focusing on training a recommender system in the source domain and deploying it in the target domain where there exist either common users or items~\cite{DBLP:conf/sigir/Yuan0KZ20,DBLP:conf/sigir/WuYCLH020,DBLP:conf/sigir/BiSYWWX20a, li2019zero}. These works are also related to the problem of recommendation for cold-start users and items, i.e., users and items with few interactions (or ratings) available during training~\cite{DBLP:conf/sigir/Hansen0SAL20,DBLP:conf/sigir/LiangXYY20,DBLP:conf/sigir/ZhuSSC20, liu2020heterogeneous}. 
There are also works~\cite{lu2020meta,DBLP:conf/kdd/DongYYXZ20} handling cold start on both user and item with meta-learning, however they cannot generalize across domains. 
In summary, prior systems are either (1) not sufficient to address our zero-shot setting where there are neither common users nor common items in the target domain or (2) unable to learn user behavior patterns that are transferable across datasets/domains.  Therefore, they are not applicable to our problem of zero-shot recommendations. 



\section{Zero-Shot Recommender Systems}


In this section we introduce our \model which is compatible with any sequential model. Without loss of generality, here we focus on NL descriptions as a possible instantiation of universal identifiers, but note that our method is general enough to use as identifiers other content information such as items' images and videos. We leave exploration for other potential modalities to future work. 

\textbf{Notation.} 
We focus on the setting of zero-shot recommendation where there are \emph{neither overlapping users nor overlapping items} between a source domain and a target domain. We assume a set $\mathcal{V}_s$ of $J_s$ items and a set $\mathcal{U}_s$ of $I_s$ users in the source domain, as well as a set $\mathcal{V}_t$ of $J_t$ items and a set $\mathcal{U}_t$ of $I_t$ users in the target domain. We let $I=I_s+I_t$ and $J=J_s+J_t$; we use $j\in \mathcal{V}_s \cup \mathcal{V}_t$ to index items and $i\in \mathcal{U}_s \cup \mathcal{U}_t$ to index users. The zero-shot setting dictates that $\mathcal{V}_s \cap \mathcal{V}_t = \emptyset$ and that $\mathcal{U}_s \cap \mathcal{U}_t = \emptyset$. We denote the collection of all users' interactions as a $3$D tensor (with necessary zero-padding) $\R\in\mathbb{R}^{I\times N_{max} \times J}$, where $N_{max}$ is the maximum number of interactions among all users. We use the subscript `$*$' to represent the collection of all elements in a certain dimension. Specifically, each user $i$ has a sequence of $N_i$ interactions (e.g., purchase history) with various items denoted as $\R_{i**} = [\R_{it*}]_{t=1}^{N_i}$, where $\R_{it*}\in \{0,1\}^{J}$ is one-hot vector denoting the $t$-th item user $i$ interacted with. 
The same user $i$ has different user embeddings at different time $t$, reflecting dynamics in user interests; here we denote as $\u_{it}\in \mathbb{R}^D$ the latent user vector when user $i$ interacts with the $t$-th item in her history, and we use $\U=[\u_{it}]_{i=1,t=1}^{I,N_{max}} \in \mathbb{R}^{I\times N_{max}\times D}$ (with necessary zero-padding) to denote the collection of user latent vectors. We denote as $\v_j\in\mathbb{R}^D$ the item latent vector and $\V=[\v_j]_{j=1}^{J}\mathbb{R}^{J\times D}$ as the collection.  
For simplicity and without loss of generality, in this paper we focus on using NL descriptions (i.e., a sequence of words) to describe items. For item $j$ we denote its NL description as $\x_j$, and the number of words as $M_j$; similar to $\V$, we let $\X=[\x_j]_{j=1}^J$. With slight notation overload on $t$, we denote as $\R^{(s)}$ and $\R^{(t)}$ the sub-tensor of $\R$ that corresponds to the source and target domains, respectively. Similarly, we also have $\U^{(s)}$, $\U^{(t)}$, $\V^{(s)}$, $\V^{(t)}$, $\X^{(s)}$, and $\X^{(t)}$.


\textbf{Problem Setup.} A model is trained using all users' interaction sequences from the source domain, i.e., $\{\R_{i**}\}_{i\in \mathcal{U}_s}$, and then deployed to recommend items for any user $\iota \in \mathcal{U}_t$ in the target domain, given user $\iota$'s previous history $\R_{\iota**}$, which can be empty. {In practice we append a dummy item at the beginning of each user session, so that during inference we could conduct recommend even for users without any history by ingesting the dummy item as context to infer the user latent vector}. In our zero-shot setting, the model is not allowed to fine-tune or retrain on any data from the target domain. 


\subsection{From Categorical Domain-Specific Item ID to Continuous Universal Item ID}
Most models in recommender systems learn item embeddings through interactions. These embeddings are indexed by \emph{categorical domain-specific item ID}, which is transductive and cannot be generalized to unseen items. 

In this paper, we propose to use item generic content information such as NL descriptions and image to produce item embeddings, which can be used as \emph{continuous universal item ID}. Since such content information is domain agnostic, the model trained on top of it can be transferable from one domain to another, therefore making zero-shot recommender systems feasible. Based on the universal item embeddings, one can then build sequential models to obtain user embeddings by aggregating embeddings of items in the user history.

Here we introduce the notion of universal embedding networks (UEN) that use continuous universal embeddings to index items (and therefore users) rather than categorical ID that is not transferable across domains. We call the UEN generating item and user universal embeddings item UEN and user UEN, respectively.

\subsection{Model Overview}\label{sec:arch}
We propose a hierarchical Bayesian model with a probabilistic encoder-decoder architecture. The encoder ingests items from user history to yield the user embedding, while decoder computes recommendation scores based on similarity between user embeddings and item embeddings. 

\textbf{Generative Process.} 
The generative process of \model (in the source domain) is as follows:
\begin{compactenum}
    \item For each item $j$:
        \begin{compactitem}
            \item Compute the item universal embedding: $\m_j = f_{e}(\x_j)$.
            \item Draw a latent item offset vector $\boldsymbol\epsilon_j \sim \mathcal{N}\left(\0, \lambda_{v}^{-1} \I_{D}\right)$.
            \item Obtain the item latent vector: 
                $
                    \v_j = \boldsymbol\epsilon_j + \m_{j}.
                $
        \end{compactitem}
    
     \item For each user $i$:
         \begin{compactitem}
         \item For each time step $t$:
         \begin{compactitem}
            \item Compute the user universal embedding: $\n_{it} = f_\textit{seq}([ \v_{i_\tau}]_{\tau=1}^{t-1})$.
            \item Draw a latent user offset vector $\boldsymbol\xi_{it} \sim \mathcal{N}\left(\0, \lambda_{u}^{-1} \I_{D}\right)$. 
            \item Obtain the latent user vector: 
                $
                    \u_{it} = \boldsymbol\xi_{it} + \n_{it}.
                $
            \item Compute recommendation score $\S_{itj}$ for each user-interaction-item tuple $(i, k, j)$, 
        $
            \S_{itj} = f_\textit{softmax}(\u_{it}^\top \v_j)
        $
        and draw the $t$-th item for user $i$: $\R_{it*} \sim Cat([\S_{itj}]_{j=1}^{J})$.
         \end{compactitem}
         \end{compactitem}
\end{compactenum}

\begin{wrapfigure}{r}{0.5\textwidth}
    
  \begin{center}
  \vskip -0.4cm
    \includegraphics[scale=0.31]{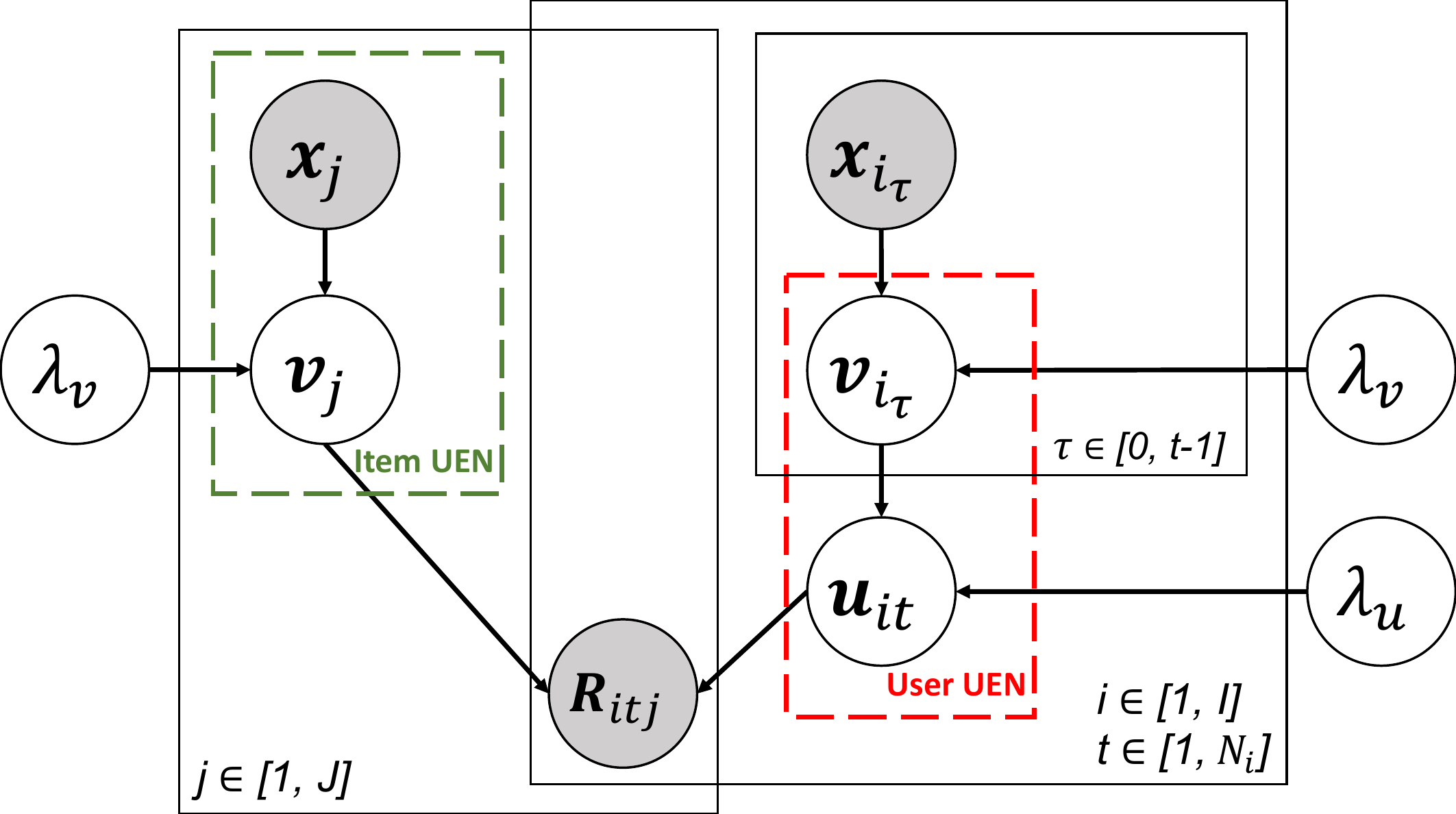}
  \end{center}
  \caption{Graphical model for \model. The item side (left) and the user side (right) share the same $\lambda_v$ and $\v$'s. The plates indicate replication. }
  \vskip -0.4cm
\end{wrapfigure}

Here $f_\textit{softmax}(\cdot)$ is the softmax function:
        $
            f_\textit{softmax}(\u_{it}^\top \v_j) = {\exp(\u_{it}^\top \v_j)}/{\sum_j \exp(\u_{it}^\top \v_j)}
        $.
$Cat(\cdot)$ is a categorical distribution. $f_{e}(\cdot)$ is item UEN (see~\secref{sec:embedding_net_item}), $f_\textit{seq}(\cdot)$ is user UEN (see~\secref{sec:embedding_net_user}). $i_\tau$ in $\v_{i_\tau}$ indexes the $\tau$-th item that user $i$ interacts with. $\lambda_{u}$ and $\lambda_{v}$ are hyperparameters. 
The latent item offset $\ep_j=\v_j-\m_j$ provides the final latent item vector $\v_j$ with the flexibility to slightly deviate from the content-based {item universal embedding} $\m_j$. {Similarly, the latent user offset $\xii_{it}=\u_{it}-\n_{it}$ provides the final latent user vector $\u_{it}$ with the flexibility to slightly deviate from the user universal embedding $\n_{it}$}. Intuitively, $\ep_j$ and $\xii_{it}$ provide domain-specific information on top of the domain-agnostic information from $\m_j$ and $\n_{it}$. 
In the target domain we will remove $\ep_j$ from $\v_j$, and $\xii_{it}$ from $\u_{it}$, which can be seen as an attempt to remove the bias learned from the source domain. 

\textbf{Training.} 
The MAP estimation in the source domain can be decomposed as following:  
\begingroup\makeatletter\def\f@size{9}\check@mathfonts
\begin{align*}
P(\U^{(s)},\V^{(s)}|\R^{(s)}, \X^{(s)}, \lambda_{u}^{-1}, \lambda_{v}^{-1}) \propto P(\R^{(s)}|\U^{(s)},\V^{(s)}) \cdot P(\U^{(s)}|\V^{(s)}, \lambda_{u}^{-1}) \cdot P(\V^{(s)}|\X^{(s)}, \lambda_{v}^{-1}).
\end{align*}
\endgroup
Maximizing the posterior probability is equivalent to minimizing the joint Negative Log-Likelihood (NLL) of $\U^{(s)}$ and $\V^{(s)}$ given $\R^{(s)}$, $\X^{(s)}$, $\lambda_{u}^{-1}$, and $\lambda_{v}^{-1}$:
\begingroup\makeatletter\def\f@size{8.5}\check@mathfonts
\begin{align}
\mathcal{L}= \sum_{i=1}^{I_s}\sum_{t=1}^{N_i}  -\log(f_\textit{softmax}(\u_{it}^\top \v_{i_t})) +
\frac{\lambda_{u}}{2} \sum_{i=1}^{I_s} \sum_{t=1}^{N_i} ||\u_{it} - f_\textit{seq}(\{ \v_{i_\tau}\}_{\tau=1}^{t-1})||_{2}^{2} + \frac{\lambda_{v}}{2} \sum_{i=1}^{J_{s}} ||\v_j - f_{e}(\x_j)||_{2}^{2}, \label{eq:overall}
\end{align}
\endgroup
where $\n_{it}=f_\textit{seq}(\{ \v_{i_\tau}\}_{\tau=1}^{t-1})$ and $\m_j=f_{e}(\x_j)$. See the Appendix for a full Bayesian treatment of \model using inference networks and generative networks~\cite{VAE}.




\textbf{Inference and Recommendation in the Target Domain.}
Once the model is trained using source-domain data, it can recommend unseen items $j\in \mathcal{V}_t$ (where $\mathcal{V}_t\cap \mathcal{V}_s=\emptyset$) for any unseen user $i\in \mathcal{U}_t$ (where $\mathcal{U}_t\cap \mathcal{U}_s=\emptyset$) from the target domain based on the approximate MAP inference below:
\begin{align*}
    p(\R^{(t)}|\X^{(t)}) &= \int p(\R^{(t)}|\U^{(t)},\V^{(t)},\X^{(t)}) p(\U^{(t)},\V^{(t)}|\X^{(t)}) d\U^{(t)} d\V^{(t)} \\
    &\approx \int p(\R^{(t)}|\U^{(t)},\V^{(t)},\X^{(t)}) \delta_{\U^{(t)}_\textit{MAP}}(\U^{(t)}) \delta_{\V^{(t)}_\textit{MAP}}(\V^{(t)}) d\U^{(t)} d\V^{(t)},
\end{align*}
where $\delta(\cdot)$ denotes a Dirac delta distribution. $\U^{(t)}_\text{MAP}$ and $\V^{(t)}_\text{MAP}$ are the MAP estimate of $\U^{(t)}$ and $\V^{(t)}$ given $\X^{(t)}$, which we approximate as:
\begin{align}
(\U^{(t)}_\textit{MAP}, \V^{(t)}_\textit{MAP}) \approx \argmax_{\U^{(t)},\V^{(t)}} p(\U^{(t)},\V^{(t)}|\X^{(t)}) = \Big(f_{\textit{seq}}(f_{e}(\X^{(t)})), f_{e}(\X^{(t)})\Big). \label{eq:inference}
\end{align}
The reason for the approximation is that \model has no access to interactions $\R^{(t)}$ in the target domain, making the cross-entropy loss in the~\eqnref{eq:overall} disappear. 
The user and item latent matrices $\U^{(t)}_\textit{MAP}, \V^{(t)}_\textit{MAP}$ in the target domain enable us to perform zero-shot recommendation by computing recommendation scores based on inner products and recommend item $\argmax_j f_\textit{softmax}(\u_{it}^\top\v_j)$.

As long as the unseen items' NL descriptions are available, \model could obtain both the unseen users' latent vectors and the unseen items' latent vectors based on the item universal embedding network. This is in contrast to previous methods that rely on catogorical domain-specific item ID, which is not transferable as unseen items have completely different ID from items in the training set. Note that \model is general enough to adapt to other data modalities such as images and videos.

\subsection{Item Universal Embedding Network:   \texorpdfstring{$\m_j=f_{e}(\x_j)$}{Lg}}\label{sec:embedding_net_item}
The purpose of the item universal embedding network, denoted as $f_{e}(\cdot)$, is to extract item embeddings that are universal across domains. The network consists of a pretrained BERT network~\cite{devlin2018bert}, denoted as $f_{\textit{BERT}}$, followed by a single-layer neural network, denoted by $f_{\textit{NN}}(\cdot)$. Formally we have
\begin{align}
\m_j = f_{e}(\x_j) = f_{\textit{NN}}(f_{\textit{BERT}}(\x_j)), \label{eq:item_uen}
\end{align}

where $\m_j$ is the universal embedding for item $j$ and $\x_j$ is the NL description for item $j$. We use the embedding for the `CLS' token from the last layer of BERT as the output of $f_{\textit{BERT}}(\x_i)$. This UEN is jointly trained with the sequential model using the objective function in~\eqnref{eq:overall}. Note that $f_{\textit{NN}}(\cdot)$ is necessary as we need to adapt the pre-trained BERT for recommendation tasks. 


\subsection{User Universal Embedding Network: \texorpdfstring{$\n_{it}=f_{\textit{seq}}([ \v_{i_\tau}]_{\tau=1}^{t-1})$}{Lg}}\label{sec:embedding_net_user}
The user UEN $f_{\textit{seq}}(\cdot)$ is built on top of the item UEN in~\eqnref{eq:item_uen}. Specifically, given user $i$'s interaction sequence until time $t$, $\R_{i**} = [\R_{i\tau*}]_{\tau=1}^{t-1}$, we first replace it the corresponding NL descriptions $\l_{i}^{(x)} = [\x_{i_\tau}]_{\tau=1}^{t-1}$, and then fed it into item UEN in~\eqnref{eq:item_uen} to obtain $\l_{i}^{(m)} =  [\m_{i_\tau}]_{\tau=1}^{t-1} = [f_{e}(\x_{i_\tau})]_{\tau=1}^{t-1}$
where item $i_t$ is user $i$'s $t$-th interaction. Each vector in the sequence, $\m_j$, is the universal embedding for item $j$. During training, we obtain item latent vector $\v_j$ based on $\m_j$ (see~\eqnref{eq:overall}), while during inference $\v_j = \m_j$ (see~\eqnref{eq:inference}). We can then treat each $\v_j$ as the input at each time step of the sequential model, which gives us the final user universal embedding $\n_{it}=f_{\textit{seq}}([ \v_{i_\tau}]_{\tau=1}^{t-1})$. 
Note that this user UEN is used during both training (\eqnref{eq:overall}) and inference (\eqnref{eq:inference}). 







\section{Experiments}
In this section, we evaluate our proposed \model against a number of in-domain and zero-shot baselines on three source-target dataset pairs, with the major goals of addressing the following questions:
\begin{compactitem}
    \item[{\bf Q1}] How accurate (effective) is \model compared to the baselines?
    \item[{\bf Q2}] If one allows training models using target-domain data, how long does it take for non zero-shot models to outperform zero-shot recommenders? 
    \item[{\bf Q3}] Does \model yield meaningful recommendations for users with similar behavioral patterns in the source domain and target domain?
\end{compactitem}



\subsection{Datasets}\label{sec:datasets}
We use three different real-world dataset pairs, one from Amazon~\cite{mcauley2015image} and two from MIND~\citep{MIND}:
\begin{compactitem}
    \item \textbf{Amazon}~\cite{mcauley2015image}: A publicly available dataset collection which contains a group of datasets in different categories with abundant item metadata such as item description, product images, etc. In our experiments, we consider two datasets: {(1) `Prime Pantry', which contains 300K interactions, 10K items, and 76K users, and (2) `Grocery and Gourmet Food', which contains 2.3M interactions, 213K items, and 739K users.} 
    \item {\textbf{MIND}~\citep{MIND}: A large-scale news recommendation dataset collected from the user click logs of Microsoft News. It includes 4-week user history and 5th-week interactions. 
    In our experiments, we simulate zero-shot learning by leave-one-out splitting on subcategories under the largest category in MIND: `sports'. We consider two pairs: {(1) `Others' to 'NFL', where `Others' contains all the other subcategories except `NFL'. `Others' contains 169K interactions, 8K items, and 57K users, while `NFL' contains 1M interactions, 11K items, and 203K users. (2) `Others' to `NCAA', where `Others' includes all the subcategories except `NCAA'. `Others' contains 797K interacions, 17K items, and 206K users, while `NCAA' contains 238K interactions, 31K items, and 56K users.} } 

\end{compactitem}

We adopted a rigorous experimental setup for zero-shot learning to ensure (1) \textbf{no overlapping} users and items and (2) \textbf{no temporal leakage}. See the Appendix for details. 
Datasets in Amazon pair are divided into training (80\%) and test (20\%) sets. For datasets in the two MIND pairs, we follow the official splitting~\citep{MIND} and use the four-week user history as the training set and interactions in the fifth week as the test set. 
Due to strong recency bias in the news recommendation~\citep{MIND}, for MIND we run experiments under two settings to measure temporal invariance of the model: (1) evaluate on interactions in the \emph{first day} of the week (\textbf{1st-day setting}), and (2) evaluate on interactions of the \emph{whole week}, i.e., the full test set (\textbf{whole-week setting}). For all datasets, 
We further split 10\% of the training set by user as validation set. 



\begin{table}
\vskip -0.6cm
\footnotesize
\caption{Zero-shot results on three dataset pairs, `Amazon Grocery and Gourmet Food' $\rightarrow$ `Amazon Prime Pantry', `Others' $\rightarrow$ `NFL', and `Others' $\rightarrow$ `NCAA'. Methods such as HRNN, HRNN-Meta, and POP are \emph{oracle methods} that are trained directly using target-domain data. N@20 and R@20 represent NDCG@20 and Recall20, respectively. The top 3 zero-shot results are shown in bold.}
\centering
  \begin{tabular}{p{30mm}p{6mm}p{6mm}p{6mm}p{6mm}p{6mm}p{6mm}p{6mm}p{6mm}p{6mm}p{6mm}}
    \toprule
    \multirow{2}{*}{\textbf{Method}} &
      \multicolumn{2}{c}{\scriptsize\textbf{\textsc{Prime Pantry}}} &
      \multicolumn{2}{c}{\scriptsize\textbf{\textsc{NCAA 1st day}}} &
      \multicolumn{2}{c}{\scriptsize\textbf{\textsc{NCAA 1 week}}} &
      \multicolumn{2}{c}{\scriptsize\textbf{\textsc{NFL 1st day}}} &
      \multicolumn{2}{c}{\scriptsize\textbf{\textsc{NFL 1 week}}} \\
      
     & \scriptsize{N@20} & \scriptsize{R@20}& \scriptsize{N@20} & \scriptsize{R@20} & \scriptsize{N@20} & \scriptsize{R@20} & \scriptsize{N@20} & \scriptsize{R@20} & \scriptsize{N@20} & \scriptsize{R@20}\\
     \midrule

    \scriptsize\textsc{HRNN (Oracle)} & $0.038$ & $0.073$ & $0.066$ & $0.139$ & $0.006$ & $0.011$ & $0.052$ & $0.118$ & $0.002$ & $0.003$ \\
    \scriptsize\textsc{HRNN-Meta (Oracle)} & $0.045$ & $0.089$ & $0.054$ & $0.120$ & $0.004$ & $0.010$ & $0.044$ & $0.112$ & $0.001$ & $0.003$ \\
    \scriptsize\textsc{GRU4Rec (Oracle)} & $0.042$ & $0.081$ & $0.062$ & $0.135$ & $0.005$ & $0.011$ & $0.046$ & $0.109$ & $0.001$ & $0.003$\\
    \scriptsize\textsc{GRU4Rec-Meta (Oracle)} & $0.044$ & $0.088$ & $0.053$ & $0.118$ & $0.004$ & $0.009$ & $0.037$ & $0.094$ & $0.001$ & $0.003$ \\
    \scriptsize\textsc{TCN (Oracle)} & $0.038$ & $0.073$ & $0.068$ & $0.141$ & $0.006$ & $0.012$ & $0.049$ & $0.114$ & $0.001$ & $0.003$ \\
    \scriptsize\textsc{TCN-Meta (Oracle)} & $0.045$ & $0.088$ & $0.054$ & $0.120$ & $0.004$ & $0.010$ & $0.044$ & $0.109$ & $0.001$ & $0.003$\\
    \scriptsize\textsc{POP (Oracle)} & $0.007$ & $0.018$ & $0.002$ & $0.005$ & $0.000$ & $0.000$ & $0.000$ & $0.000$ & $0.000$ & $0.000$ \\ \midrule
    \scriptsize\textsc{EMB-KNN (Baseline)} & $0.024$ & $0.042$ & $0.016$ & $0.026$ & $0.005$ & $0.011$ & $\textbf{0.010}$ & $\textbf{0.018}$ & $0.001$ & $0.003$\\
    \scriptsize\textsc{Random (Baseline)} & $0.001$ & $0.002$ & $0.002$ & $0.006$ & $0.002$ & $0.006$ & $0.001$ & $0.002$ & $0.001$ & $0.002$\\
    \scriptsize\textsc{\model-H (Ours)} & $\textbf{0.027}$ & $\textbf{0.052}$ & $\textbf{0.027}$ & $\textbf{0.063}$ & $\textbf{0.011}$ & $\textbf{0.036}$ & $0.007$ & $0.013$ & $\textbf{0.015}$ & $\textbf{0.043}$\\ 
    \scriptsize\textsc{\model-G (Ours)} & $\textbf{0.026}$ & $\textbf{0.050}$ & $\textbf{0.030}$ & $\textbf{0.070}$ & $\textbf{0.014}$ & $\textbf{0.040}$ & $\textbf{0.008}$ & $\textbf{0.013}$ & $\textbf{0.018}$ & $\textbf{0.058}$\\ 
    \scriptsize\textsc{\model-T (Ours)} & $\textbf{0.026}$ & $\textbf{0.050}$ & $\textbf{0.023}$ & $\textbf{0.056}$ & $\textbf{0.011}$ & $\textbf{0.035}$ & $\textbf{0.009}$ & $\textbf{0.017}$ & $\textbf{0.018}$ & $\textbf{0.054}$\\ \bottomrule
  \end{tabular}
\centering
\label{table:zsl_results}
\vskip -0.6cm
\end{table}

\subsection{Baselines and \model Variants}
To demonstrate the effectiveness of our model, we {compare \model against} two groups of baselines: in-domain methods and zero-shot methods. 

\textbf{In-Domain Methods.} We compare variants of our model \model against a variety of state-of-the-art session-based recommendation models including \textbf{GRU4Rec}~\cite{GRU4Rec}, \textbf{TCN}~\cite{TCN}, and \textbf{HRNN}~\cite{HRNN}. We also consider their extensions, \textbf{HRNN-Meta}, \textbf{GRU4Rec-Meta}, and \textbf{TCN-Meta}, which 
use items' NL description embeddings to replace item ID hidden embeddings. Besides aforementioned sophisticated models, we introduce \textbf{POP} which is a simple baseline conducting recommendation only based on item global popularity. It can be a strong baseline in certain domains. All the above $7$ methods are trained directly on target-domain data and therefore are considered \emph{`oracle'} methods. 

\textbf{Zero-Shot Methods.} Since no previous work has been done on this thread, we consider two intuitive zero-shot models (1) \textbf{EmbeddingKNN}: {a K-nearest-neighbors algorithm based on inner product between the user embedding and item embedding. The item embedding is the BERT embedding from NL description, while the user embedding is an average over embeddings of the user's interacted items}, and (2) \textbf{Random}: recommending items by random selection from the whole item catalogue without replacement.

\textbf{\model Variants.} We evaluate three variants of our \model, including \textbf{\model-G}, \textbf{\model-T}, and \textbf{\model-H} which use GRU4Rec, TCN and HRNN as base models, respectively.

\begin{figure*}
\vskip -0.7cm
    \centering
    \subfloat[NDCG@20]{{\includegraphics[width=0.235\textwidth]{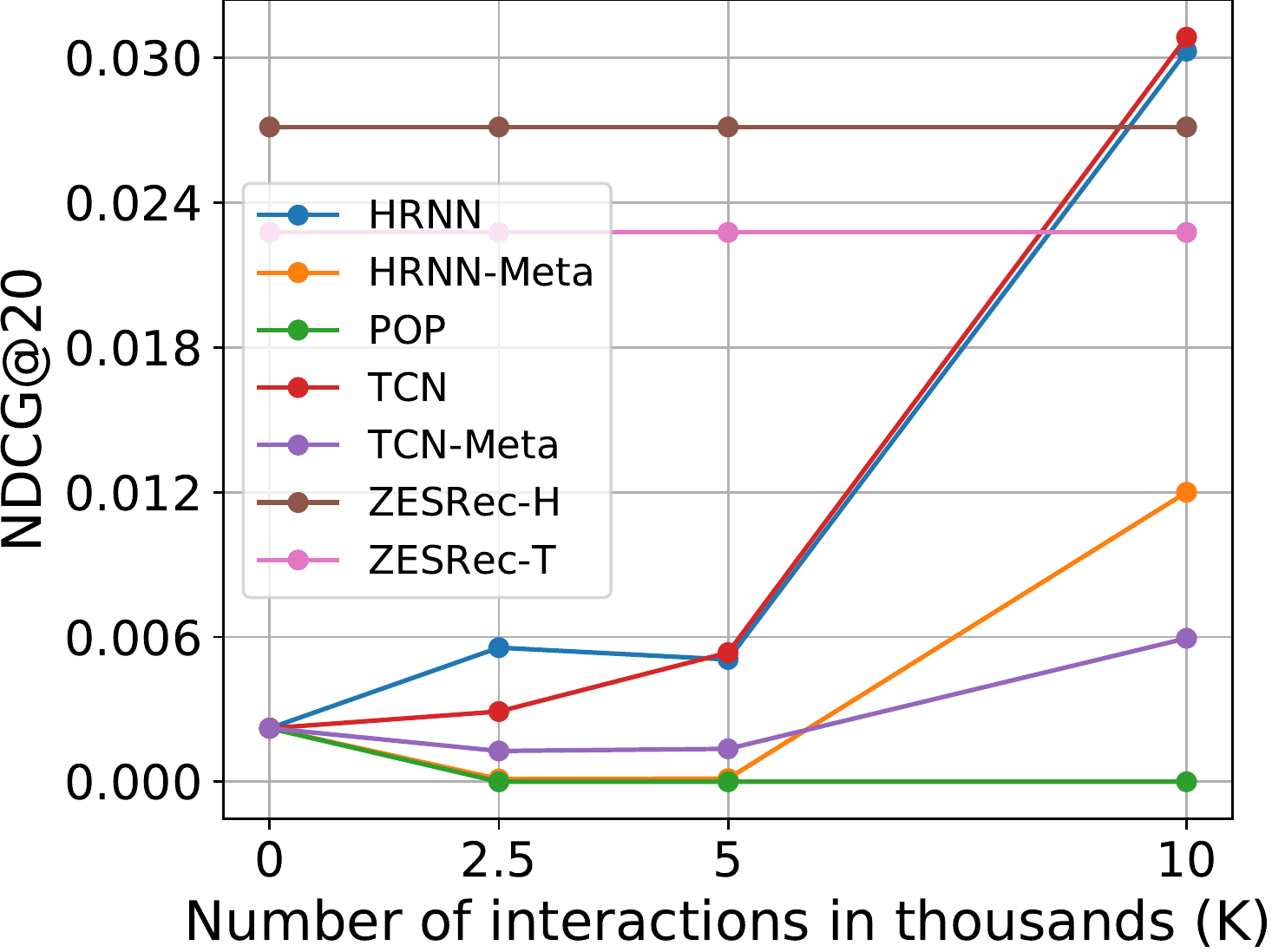} } \label{fig:mind_ndcg}}
    \subfloat[Recall@20]{{\includegraphics[width=0.235\textwidth]{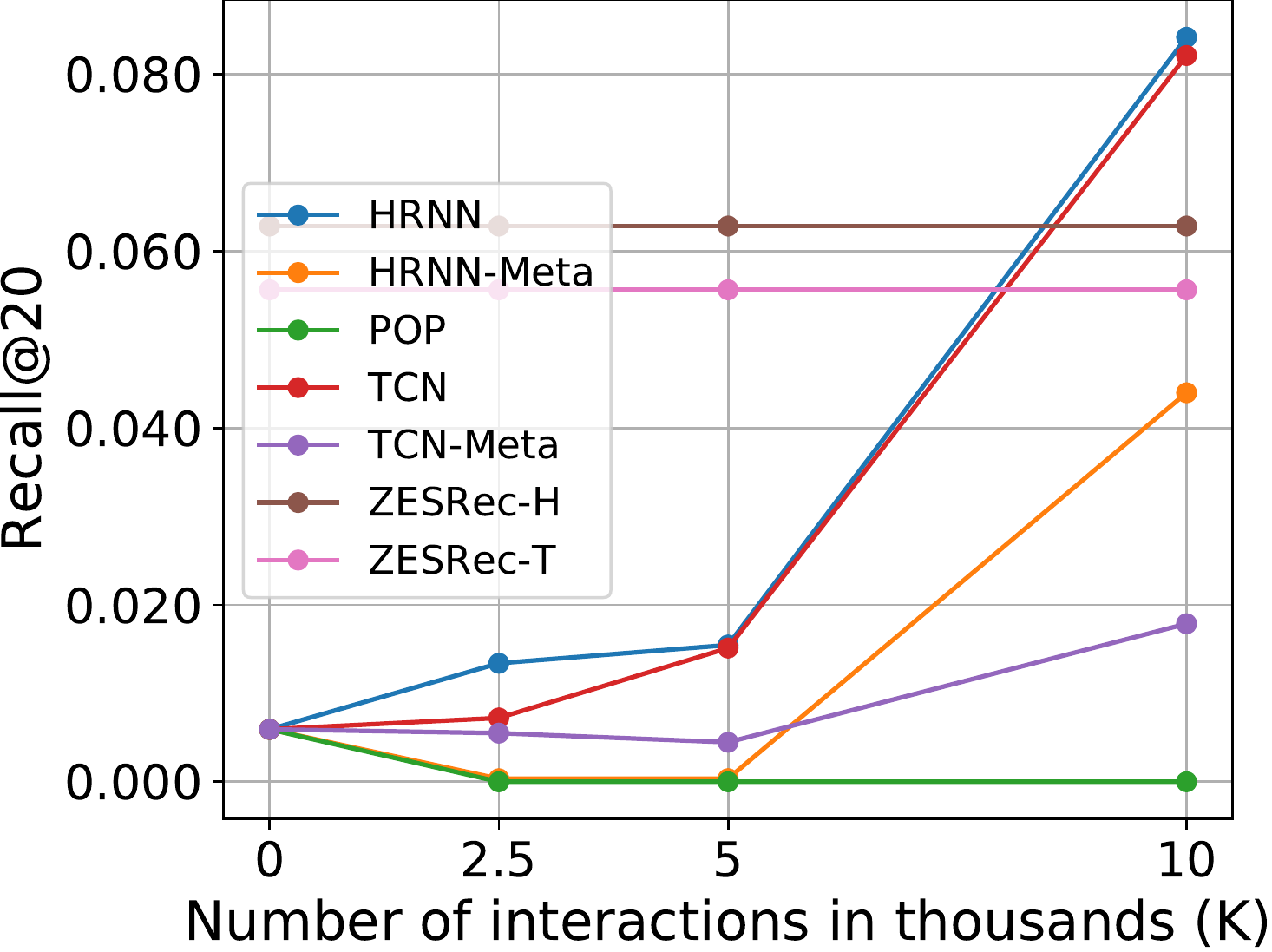} } \label{fig:mind_recall}}
    \subfloat[NDCG@20]{{\includegraphics[width=0.23\textwidth]{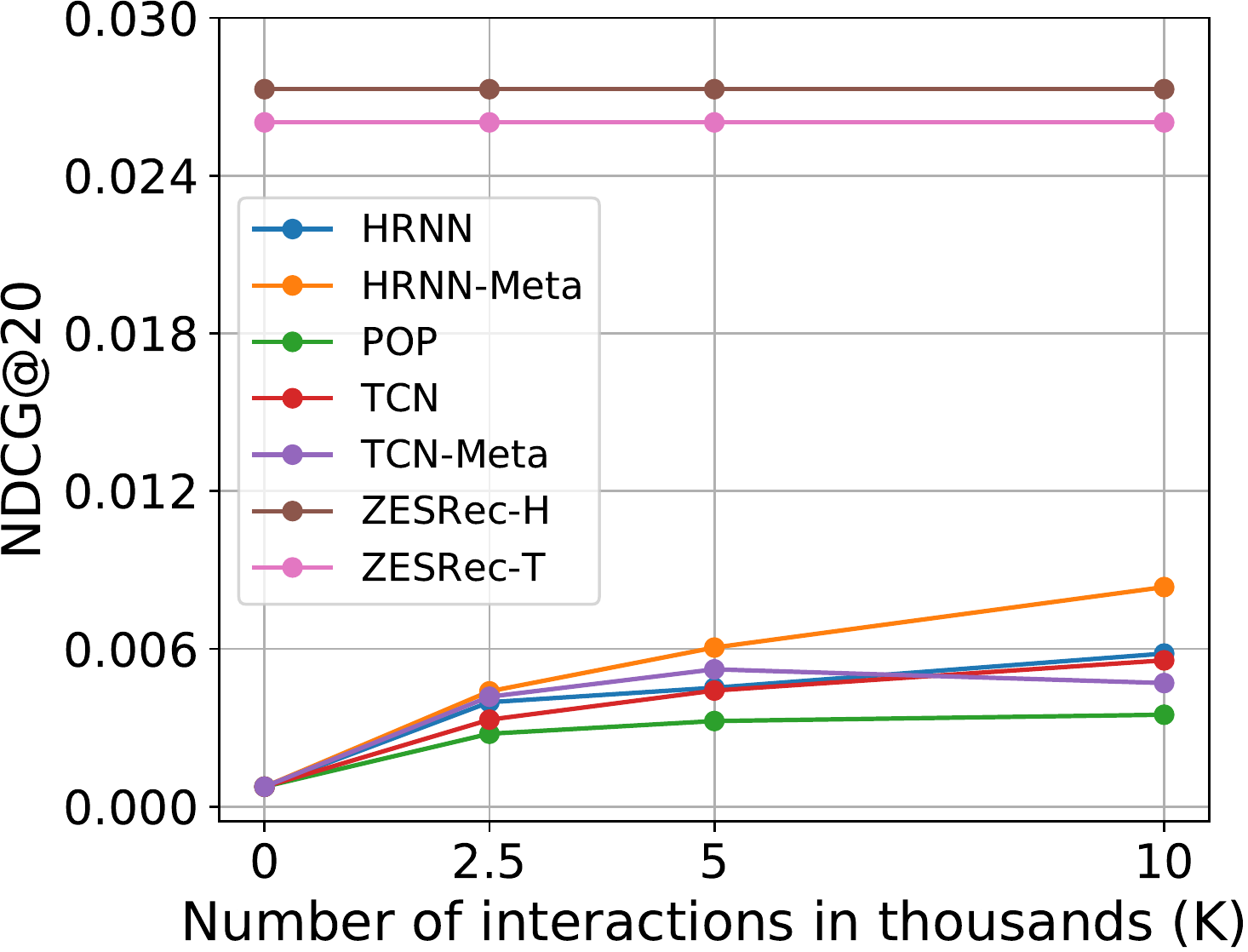} } \label{fig:pp_ndcg}}
    \subfloat[Recall@20.]{{\includegraphics[width=0.23\textwidth]{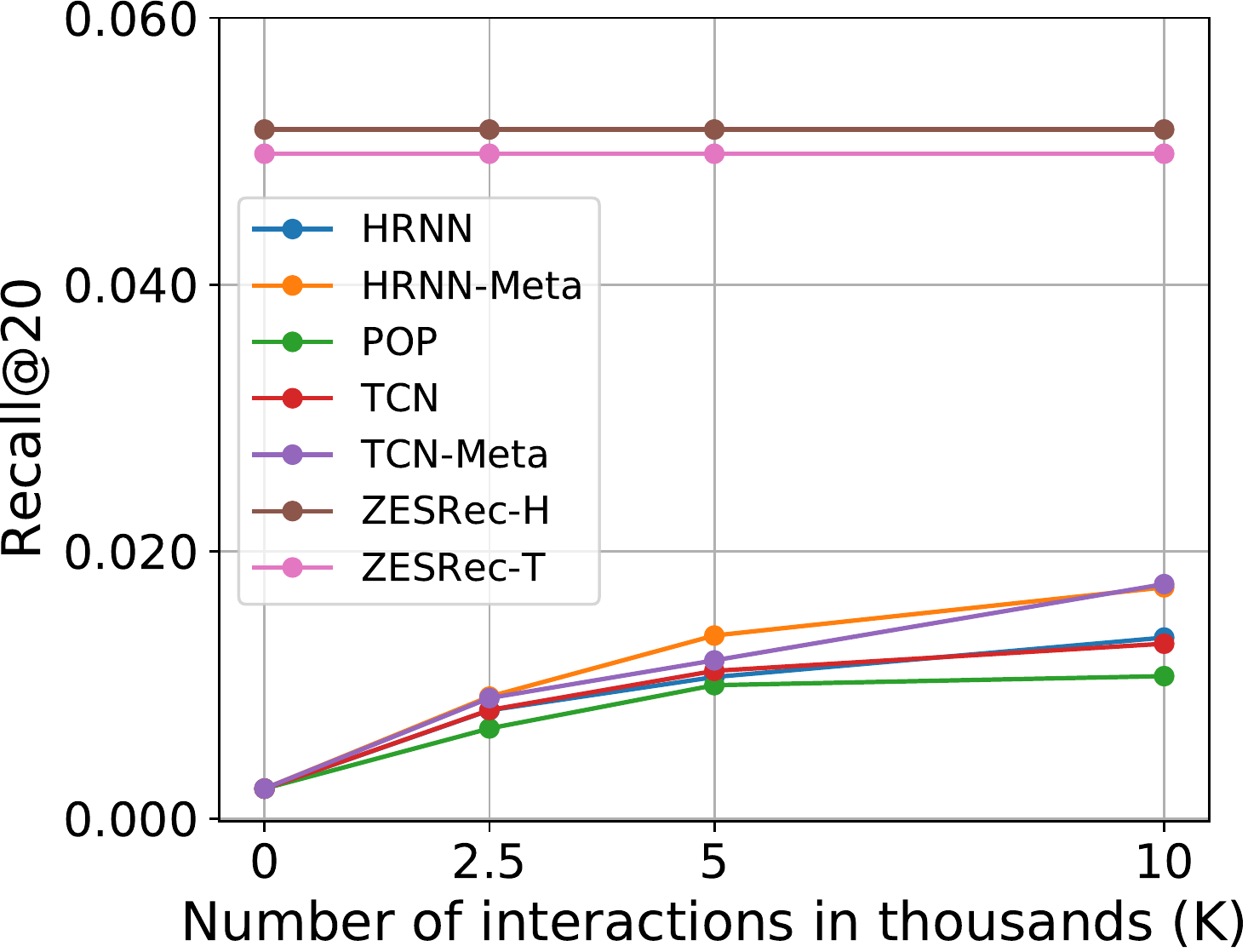} } \label{fig:pp_recall}}
\vskip -0.3cm
    \caption{Incremental training results for baselines using target domain data compared to \model using \emph{no data} on MIND-NCAA (left two) and Amazon Prime Pantry (right two). To prevent clutter, we only show results for TCN-based and HRNN-based models, since HRNN is an advanced version of GRU4Rec. Results show that even without using target-domain data, \model can still outperform models trained directly using target-domain data for substantial amount of time.
    }
    \label{fig:incremental_exp}
    \vskip -0.6cm
\end{figure*}

\subsection{Implementation Details}\label{sec:imp_details}
We adopted pre-trained \emph{google/bert\_uncased\_L-12\_H-768\_A-12} BERT model from Huggingface~\cite{HuggingFace} to process item description and generate item embedding. The dimension of BERT embedding is 768. For the model architecture, we use BERT embedding as input to a single-layer neural netwrok (i.e., $f_\textit{NN}(\cdot)$). The output dimension for the NN is set to $D$ (see the Appendix for more details). 

\subsection{Experiment Setup and Evaluation Metrics}\label{sec:exp_setup}
We conducted three sets of experiments to answer questions proposed at the begining part of the experiment section, specifically:

\textbf{Zero-Shot Experiments.} We trained in-domain baselines on target domain training set, while our \model is trained on source domain. All models are tested on the testing set of the target domain for an apples to apples comparison.

\textbf{Incremental Training Experiments.} To measure how long it takes for non-zero-shot models 
to outperform zero-shot recommenders, we conducted incremental training experiments on in-domain base models GRU4Rec, TCN, HRNN as well as GRU4Rec-Meta, TCN-Meta, HRNN-Meta. Note that the variants of our \model are NOT retrained on target domain. 
It is also \emph{inevitable} that non-zero-shot models eventually outperform \model because \model does not have access to target-domain data. 
Due to space constraint, we only reported results for incremental training experiments on Amazon pair and MIND NCAA pair under the 1st-day setting. Importantly, under the whole-week setting for the MIND dataset, the ZESRec variants already significantly outperform in-domain competitors which are trained on the full target training set; therefore it is meaningless to compare incremental training results under this setting, as in-domain competitors will never beat ZESRec variants. 

For all the source-target dataset pairs, we group the interactions by user and sort interactions within each user based on interaction timestamps. We randomly select users until we get enough interactions and build three datasets which contain 2.5K, 5K, and 10K interactions, respectively.


\begin{figure*}[!t]
\vskip -0.7cm
\centering
\includegraphics[width=0.97\textwidth]{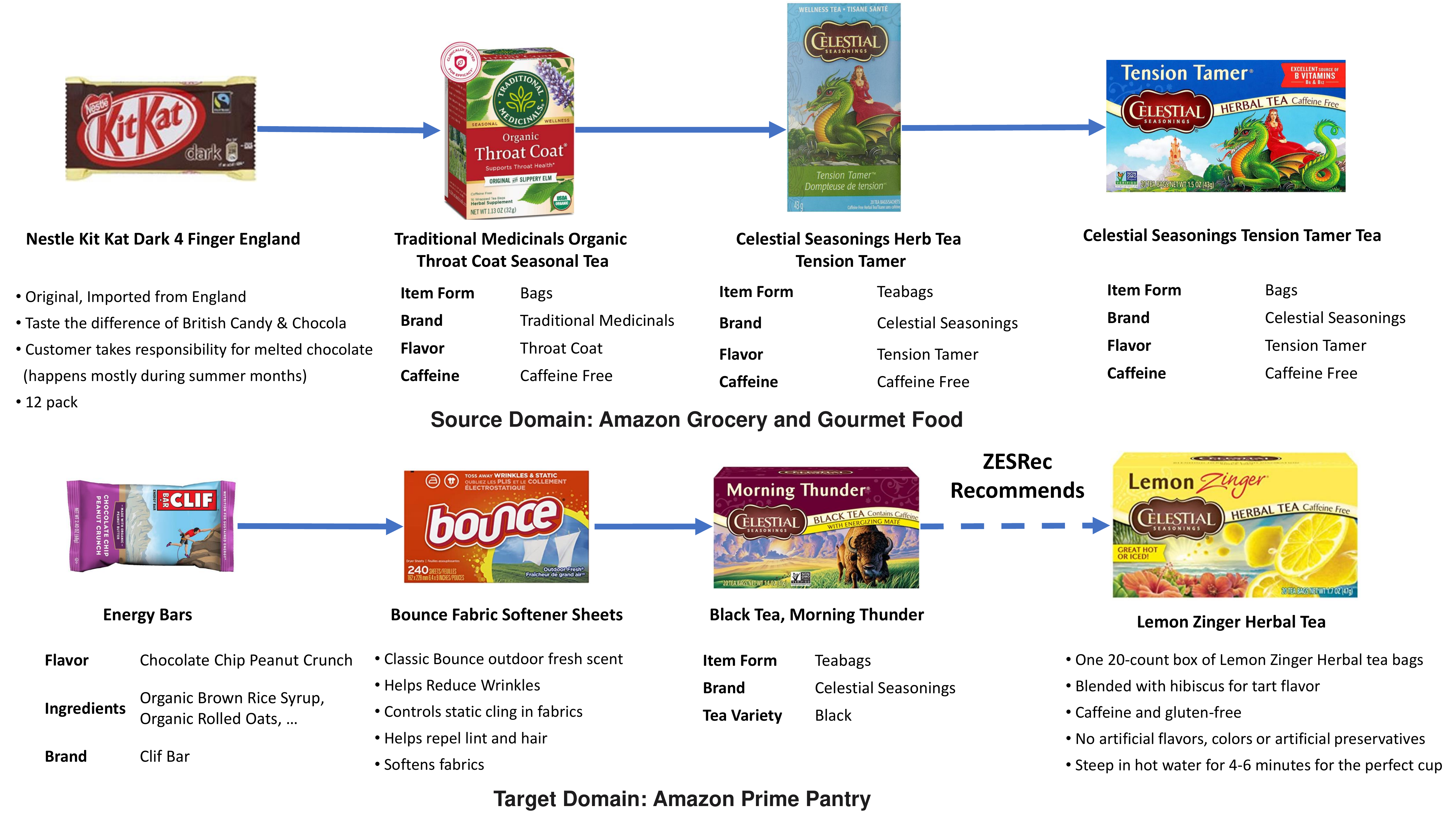}
\vskip -0.3cm
\caption{Case Study 1. The purchase history of a user in the source domain (\emph{top}) and the purchase history of an unseen user in the target domain, where all items are unseen during training (\emph{bottom}). We select two users with similar universal embeddings according to~\secref{sec:exp_setup}. This case study demonstrates \model can learn the user behavioral pattern that `users who bought sugary snacks and tea tend to buy caffeine-free herbal tea later'. }
\label{fig:case_study_1}
\vskip -0.7cm
\end{figure*}

\textbf{Case Studies.} 
To gain more insight what \model learns, we perform several case studies. Specifically, we randomly select users from the test set of the target domain Amazon Prime Pantry (where we evaluate \model) and only keep users for whom \model correctly predicts the 6-th items in the sequence given the first 5 items as context, as we want to focus on sequences where \model works. We use these users as queries to find users with similar behavioral patterns from the source domain Amazon Grocery and Gourmet Food (where we train \model) based on user embeddings from \model. User embeddings are generated based on the first 5 items of the sequence. 


\textbf{Evaluation Protocol.} For evaluation, we adopted Recall (R@20) and the ranking metric Normalized Discounted Cumulative Gain (NDCG)~\cite{NDCG} (N@20). We removed all the repetitive interactions (e.g., user A clicked item B two times in a row) to only focus on evaluating the model's capability of capturing the transition between user history to the next item. 

\subsection{Experimental Results}

\textbf{Zero-Shot Experiments.} The experimental results on three dataset pairs are in Table~\ref{table:zsl_results}. {Overall, our \model outperforms zero-shot baselines Embedding-KNN and Random by a large margin in most cases; it can also achieve performance comparable to in-domain baselines. }

On the Amazon pair, \model beats POP and achieves comparable performance with HRNN-Meta and HRNN-Interactions, suggesting the existence of shared recommendation patterns in two domains. {For the two MIND pairs, we made the following observations: (1) Under the whole-week setting, \model consistently outperforms HRNN-Meta and HRNN-Interactions by a large margin. Comparing the first-day results and the whole-week results, it is obvious that the in-domain models overfit the history and fail to generalize well on latest user-item interactions. 
This reflects strong recency bias in news recommendation. In contrast, our zero-shot learning naturally comes with an inductive bias to only model the transition from user history to the next item. (2) Under the first-day setting, \model still achieves reasonable performance comparing with in-domain models on both pairs.} 
{(3) Surprisingly in the MIND NFL pair, the ZESRec variants perform well under the whole-week setting even when the source domain contains much fewer interactions than the target domain (169K VS 1M).}


\textbf{Incremental Training Experiments.} 
Incremental training results are in ~\figref{fig:incremental_exp}. Overall, almost all the in-domain baselines are not able to outperform \model by retraining on at most 10K interactions in target domain, which shows the vast importance of conducting zero-shot learning in recommender domain. 
For new business operating an early-stage RecSys, it's hard to train a good RecSys with limited interactions. This is a chicken-and-egg problem, as training good RecSys requires sufficient interactions, while in turn, collecting sufficient interactions requires a satisfactory RecSys to attract users. 
Therefore the first 10K interactions are crucial to get the RecSys started. 

\begin{figure*}
\vskip -0.3cm
\centering
\includegraphics[width=0.97\textwidth]{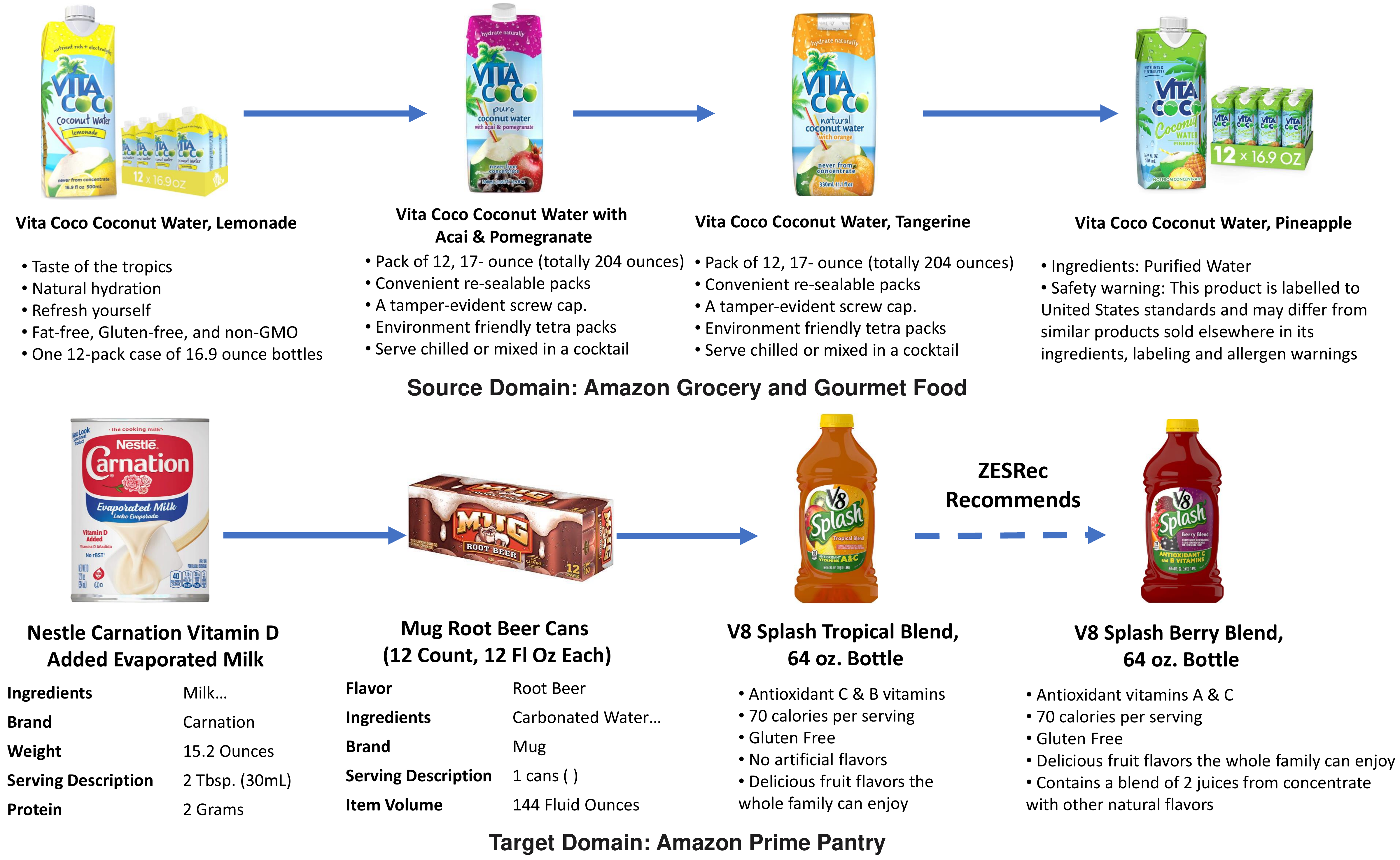}
\vskip -0.3cm
\caption{Case Study 2. The purchase history of a user in the source domain (\emph{top}) and the purchase history of an unseen user in the target domain, where all items are unseen during training (\emph{bottom}). We select two users with similar universal embeddings according to~\secref{sec:exp_setup}. This case study demonstrates \model can learn the user behavioral pattern that `if users bought snacks or drinks that they like, they may later purchase similar snacks or drinks with different flavors'. }
\label{fig:case_study_2}
\vskip -0.6cm
\end{figure*}

\subsection{Case Studies}\label{sec:case}
To gain more insight, we randomly select pairs of user purchase histories from the source and target domains {in the Amazon pair} as case studies according to the procedure in~\secref{sec:exp_setup}. The goal is to demonstrate our ZSR could learn relevant dynamics of users' purchase history from the source domain and successfully recommend unseen products to an unseen user in the target domain.

\figref{fig:case_study_1} shows the purchase history of a user in the source domain (\emph{top}), `Amazon Grocery and Gourmet Food', and the purchase history of an unseen user in the target domain (\emph{bottom}), `Amazon Prime Pantry', where all items are unseen during training. The user in the source domain bought `Tension Tamer Tea', which is a type of herbal tea, after buying some sugary snacks (KitKat) and other tea. Such a pattern is captured by \model, which then recommended `Lemon Zinger Herbal Tea' to an unseen user after she bought some sugary snacks (`Energy Bars from Clif Bar') and some black tea. This case study demonstrates \model can learn the user behavioral pattern that `users who bought sugary snacks and tea tend to buy caffeine-free herbal tea later'.

More interestingly, another case study in~\figref{fig:case_study_2} demonstrates that \model can learn the user behavioral pattern that `if users bought snacks or drinks that they like, they may later purchase similar snacks or drinks with different flavors'. Specifically, in the source domain, the user purchased `Vita Coconut Water' with four different flavors; such a pattern is captured by \model. Later in the target domain, an unseen user purchase `V8 Splash' with a tropical flavor, \model then successfully recommends `V8 Splash' with a berry flavor to the user.




\section{Conclusion}
In this paper, we identify the problem of zero-shot recommender systems where a model is trained in the source domain and deployed in a target domain where all the users and items are unseen during training. We propose \model as the first general framework for addressing this problem. We introduce the notion of universal continuous identifiers leveraging the fact that item ID can be grounded in natural-language descriptions. We provide empirical results, both quantitatively and qualitatively, to demonstrate the effectiveness of our proposed \model and verify that \model can successfully learn user behavioral patterns that generalize across datasets (domains). Future work includes exploring other modalities, e.g., images and videos, as alternative universal identifiers. It would also be interesting to investigate the interpretability provided by the pretrained BERT model and to incorporate additional auxiliary information to further improve the zero-shot performance.

\bibliography{iclr2022_conference}

\begin{thebibliography}{10}

\bibitem{TCN}
{\sc Bai, S., Kolter, J.~Z., and Koltun, V.}
\newblock An empirical evaluation of generic convolutional and recurrent
  networks for sequence modeling.
\newblock {\em CoRR abs/1803.01271\/} (2018).

\bibitem{DBLP:conf/sigir/BiSYWWX20a}
{\sc Bi, Y., Song, L., Yao, M., Wu, Z., Wang, J., and Xiao, J.}
\newblock A heterogeneous information network based cross domain insurance
  recommendation system for cold start users.
\newblock In {\em SIGIR\/} (2020), pp.~2211--2220.

\bibitem{chen2019top}
{\sc Chen, M., Beutel, A., Covington, P., Jain, S., Belletti, F., and Chi,
  E.~H.}
\newblock Top-k off-policy correction for a {REINFORCE} recommender system.
\newblock In {\em WSDM\/} (2019), pp.~456--464.

\bibitem{GRU}
{\sc Cho, K., van Merrienboer, B., Gulcehre, C., Bahdanau, D., Bougares, F.,
  Schwenk, H., and Bengio, Y.}
\newblock Learning phrase representations using {RNN} encoder-decoder for
  statistical machine translation.
\newblock In {\em EMNLP\/} (2014), pp.~1724--1734.

\bibitem{devlin2018bert}
{\sc Devlin, J., Chang, M.-W., Lee, K., and Toutanova, K.}
\newblock Bert: Pre-training of deep bidirectional transformers for language
  understanding.
\newblock {\em arXiv preprint arXiv:1810.04805\/} (2018).

\bibitem{DBLP:conf/kdd/DongYYXZ20}
{\sc Dong, M., Yuan, F., Yao, L., Xu, X., and Zhu, L.}
\newblock {MAMO:} memory-augmented meta-optimization for cold-start
  recommendation.
\newblock In {\em KDD\/} (2020), pp.~688--697.

\bibitem{fang2019deep}
{\sc Fang, H., Zhang, D., Shu, Y., and Guo, G.}
\newblock Deep learning for sequential recommendation: Algorithms, influential
  factors, and evaluations.
\newblock {\em arXiv preprint arXiv:1905.01997\/} (2019).

\bibitem{DBLP:conf/sigir/Hansen0SAL20}
{\sc Hansen, C., Hansen, C., Simonsen, J.~G., Alstrup, S., and Lioma, C.}
\newblock Content-aware neural hashing for cold-start recommendation.
\newblock In {\em SIGIR\/} (2020), pp.~971--980.

\bibitem{GRU4Rec}
{\sc Hidasi, B., Karatzoglou, A., Baltrunas, L., and Tikk, D.}
\newblock Session-based recommendations with recurrent neural networks.
\newblock {\em arXiv preprint arXiv:1511.06939\/} (2015).

\bibitem{MF}
{\sc Hu, Y., Koren, Y., and Volinsky, C.}
\newblock Collaborative filtering for implicit feedback datasets.
\newblock In {\em ICDM\/} (2008), pp.~263--272.

\bibitem{SASRec}
{\sc Kang, W.-C., and McAuley, J.}
\newblock Self-attentive sequential recommendation.
\newblock In {\em 2018 IEEE International Conference on Data Mining (ICDM)\/}
  (2018), IEEE, pp.~197--206.

\bibitem{VAE}
{\sc Kingma, D.~P., and Welling, M.}
\newblock Auto-encoding variational bayes.
\newblock In {\em ICLR\/} (2014).

\bibitem{li2019zero}
{\sc Li, J., Jing, M., Lu, K., Zhu, L., Yang, Y., and Huang, Z.}
\newblock From zero-shot learning to cold-start recommendation.
\newblock In {\em Proceedings of the AAAI Conference on Artificial
  Intelligence\/} (2019), vol.~33, pp.~4189--4196.

\bibitem{NARM}
{\sc Li, J., Ren, P., Chen, Z., Ren, Z., Lian, T., and Ma, J.}
\newblock Neural attentive session-based recommendation.
\newblock In {\em Proceedings of the 2017 ACM on Conference on Information and
  Knowledge Management\/} (2017), pp.~1419--1428.

\bibitem{li2017neural}
{\sc Li, J., Ren, P., Chen, Z., Ren, Z., Lian, T., and Ma, J.}
\newblock Neural attentive session-based recommendation.
\newblock In {\em CIKM\/} (2017), pp.~1419--1428.

\bibitem{CVAE}
{\sc Li, X., and She, J.}
\newblock Collaborative variational autoencoder for recommender systems.
\newblock In {\em KDD\/} (2017), pp.~305--314.

\bibitem{DBLP:conf/sigir/LiangXYY20}
{\sc Liang, T., Xia, C., Yin, Y., and Yu, P.~S.}
\newblock Joint training capsule network for cold start recommendation.
\newblock In {\em SIGIR\/} (2020), pp.~1769--1772.

\bibitem{STAMP}
{\sc Liu, Q., Zeng, Y., Mokhosi, R., and Zhang, H.}
\newblock Stamp: short-term attention/memory priority model for session-based
  recommendation.
\newblock In {\em Proceedings of the 24th ACM SIGKDD International Conference
  on Knowledge Discovery \& Data Mining\/} (2018), pp.~1831--1839.

\bibitem{liu2018stamp}
{\sc Liu, Q., Zeng, Y., Mokhosi, R., and Zhang, H.}
\newblock Stamp: short-term attention/memory priority model for session-based
  recommendation.
\newblock In {\em KDD\/} (2018), pp.~1831--1839.

\bibitem{liu2020heterogeneous}
{\sc Liu, S., Ounis, I., Macdonald, C., and Meng, Z.}
\newblock A heterogeneous graph neural model for cold-start recommendation.
\newblock In {\em Proceedings of the 43rd International ACM SIGIR Conference on
  Research and Development in Information Retrieval\/} (2020), pp.~2029--2032.

\bibitem{lu2020meta}
{\sc Lu, Y., Fang, Y., and Shi, C.}
\newblock Meta-learning on heterogeneous information networks for cold-start
  recommendation.
\newblock In {\em Proceedings of the 26th ACM SIGKDD International Conference
  on Knowledge Discovery \& Data Mining\/} (2020), pp.~1563--1573.

\bibitem{HRNN}
{\sc Ma, Y., Narayanaswamy, B., Lin, H., and Ding, H.}
\newblock Temporal-contextual recommendation in real-time.
\newblock In {\em Proceedings of the 26th ACM SIGKDD International Conference
  on Knowledge Discovery \& Data Mining\/} (2020), pp.~2291--2299.

\bibitem{mcauley2015image}
{\sc McAuley, J., Targett, C., Shi, Q., and Van Den~Hengel, A.}
\newblock Image-based recommendations on styles and substitutes.
\newblock In {\em Proceedings of the 38th international ACM SIGIR conference on
  research and development in information retrieval\/} (2015), pp.~43--52.

\bibitem{HGRU}
{\sc Quadrana, M., Karatzoglou, A., Hidasi, B., and Cremonesi, P.}
\newblock Personalizing session-based recommendations with hierarchical
  recurrent neural networks.
\newblock In {\em RecSys\/} (2017), pp.~130--137.

\bibitem{PMF}
{\sc Salakhutdinov, R., and Mnih, A.}
\newblock Probabilistic matrix factorization.
\newblock In {\em NIPS\/} (2007), pp.~1257--1264.

\bibitem{RBM4CF}
{\sc Salakhutdinov, R., Mnih, A., and Hinton, G.~E.}
\newblock Restricted boltzmann machines for collaborative filtering.
\newblock In {\em ICML\/} (2007), vol.~227, pp.~791--798.

\bibitem{NDCG}
{\sc Shani, G., and Gunawardana, A.}
\newblock Evaluating recommendation systems.
\newblock In {\em Recommender systems handbook}. Springer, 2011, pp.~257--297.

\bibitem{tang2019towards}
{\sc Tang, J., Belletti, F., Jain, S., Chen, M., Beutel, A., Xu, C., and
  H.~Chi, E.}
\newblock Towards neural mixture recommender for long range dependent user
  sequences.
\newblock In {\em WWW\/} (2019), pp.~1782--1793.

\bibitem{deepmusic}
{\sc van~den Oord, A., Dieleman, S., and Schrauwen, B.}
\newblock Deep content-based music recommendation.
\newblock In {\em NIPS\/} (2013), pp.~2643--2651.

\bibitem{CDL}
{\sc Wang, H., Wang, N., and Yeung, D.}
\newblock Collaborative deep learning for recommender systems.
\newblock In {\em KDD\/} (2015), pp.~1235--1244.

\bibitem{HuggingFace}
{\sc Wolf, T., Debut, L., Sanh, V., Chaumond, J., Delangue, C., Moi, A.,
  Cistac, P., Rault, T., Louf, R., Funtowicz, M., Davison, J., Shleifer, S.,
  von Platen, P., Ma, C., Jernite, Y., Plu, J., Xu, C., Scao, T.~L., Gugger,
  S., Drame, M., Lhoest, Q., and Rush, A.~M.}
\newblock Transformers: State-of-the-art natural language processing.
\newblock In {\em Proceedings of the 2020 Conference on Empirical Methods in
  Natural Language Processing: System Demonstrations\/} (Online, Oct. 2020),
  Association for Computational Linguistics, pp.~38--45.

\bibitem{MIND}
{\sc Wu, F., Qiao, Y., Chen, J.-H., Wu, C., Qi, T., Lian, J., Liu, D., Xie, X.,
  Gao, J., Wu, W., et~al.}
\newblock Mind: A large-scale dataset for news recommendation.
\newblock In {\em Proceedings of the 58th Annual Meeting of the Association for
  Computational Linguistics\/} (2020), pp.~3597--3606.

\bibitem{DBLP:conf/sigir/WuYCLH020}
{\sc Wu, L., Yang, Y., Chen, L., Lian, D., Hong, R., and Wang, M.}
\newblock Learning to transfer graph embeddings for inductive graph based
  recommendation.
\newblock In {\em SIGIR\/} (2020), pp.~1211--1220.

\bibitem{wu2019session}
{\sc Wu, S., Tang, Y., Zhu, Y., Wang, L., Xie, X., and Tan, T.}
\newblock Session-based recommendation with graph neural networks.
\newblock In {\em AAAI\/} (2019), vol.~33, pp.~346--353.

\bibitem{DBLP:conf/sigir/Yuan0KZ20}
{\sc Yuan, F., He, X., Karatzoglou, A., and Zhang, L.}
\newblock Parameter-efficient transfer from sequential behaviors for user
  modeling and recommendation.
\newblock In {\em SIGIR\/} (2020), pp.~1469--1478.

\bibitem{DBLP:conf/sigir/ZhuSSC20}
{\sc Zhu, Z., Sefati, S., Saadatpanah, P., and Caverlee, J.}
\newblock Recommendation for new users and new items via randomized training
  and mixture-of-experts transformation.
\newblock In {\em SIGIR\/} (2020), pp.~1121--1130.

\end{thebibliography}
\setcitestyle{square}
\bibliographystyle{acm}

\end{document}